\documentclass[11pt,a4paper]{article}
\usepackage[hyperref]{naaclhlt2019}
\usepackage{times}
\usepackage{latexsym}
\usepackage{graphicx}
\usepackage{booktabs}
\graphicspath{ {images/} }

\usepackage{url}
\usepackage{verbatim}

\aclfinalcopy 

\title{Learning Multilingual Word Embeddings Using Image-Text Data}

\author{Karan Singhal \\
  Stanford University \\
  {\tt ksinghal@cs.stanford.edu} \\\AND
  Karthik Raman \\
  Google AI \\
  {\tt karthikraman@google.com} \\\And
  Balder ten Cate \\
  Google AI \\
  {\tt balder@google.com} \\}

\date{}

\begin{document}
\maketitle
\begin{abstract}
  There has been significant interest recently in learning multilingual word embeddings -- in which semantically similar words across languages have similar embeddings. State-of-the-art approaches have relied on expensive labeled data, which is unavailable for low-resource languages, or have involved post-hoc unification of monolingual embeddings. In the present paper, we investigate the efficacy of multilingual embeddings learned from weakly-supervised image-text data. In particular, we propose methods for learning multilingual embeddings using image-text data, by enforcing similarity between the representations of the image and that of the text. Our experiments reveal that even without using any expensive labeled data, a bag-of-words-based embedding model trained on image-text data achieves performance comparable to the state-of-the-art on crosslingual semantic similarity tasks.
\end{abstract}

\section{Introduction}
\label{Introduction}
Recent advances in learning distributed representations for words (\emph{i.e.,} word embeddings) have resulted in improvements across numerous natural language understanding tasks \cite{word2vec, glove}. These methods use unlabeled text corpora to model the semantic content of words using their co-occurring context words. Key to this is the observation that semantically similar words have similar contexts \cite{sahlgren2008distributional}, thus leading to similar word embeddings. A limitation of these word embedding approaches is that they only produce \emph{monolingual embeddings}. This is because word co-occurrences are very likely to be limited to being within language rather than across language in text corpora. Hence semantically similar words across languages are unlikely to have similar word embeddings. 

To remedy this, there has been recent work on learning \emph{multilingual word embeddings}, in which semantically similar words within \emph{and} across languages have similar word embeddings \cite{ruder17-crosslingsurvey}. Multilingual embeddings are not just interesting as an interlingua between multiple languages; they are useful in many downstream applications. For example, one application of multilingual embeddings is to find semantically similar words and phrases across languages \cite{Ammar}. Another use of multilingual embeddings is in enabling zero-shot learning on unseen languages, just as monolingual word embeddings enable predictions on unseen words \cite{artetxecrosslingualtransfer}. In other words, a classifier using pretrained multilingual word embeddings can generalize to other languages even if training data is only in English. Interestingly, multilingual embeddings have also been shown to improve monolingual task performance \cite{Faruqui, kiela2014improving}. 

Consequently, multilingual embeddings can be very 
useful for low-resource languages -- they allow us to overcome the scarcity of data in these languages. However, as detailed in Section \ref{Related}, most work on learning multilingual word embeddings so far has heavily relied on the availability of expensive resources such as word-aligned / sentence-aligned parallel corpora or bilingual lexicons. Unfortunately, this data can be prohibitively expensive to collect for many languages. Furthermore even for languages with such data available, the coverage of the data is a limiting factor that restricts how much of the semantic space can be aligned across languages. Overcoming this data bottleneck is a key contribution of our work.

We investigate the use of cheaply available, weakly-supervised image-text data for learning multilingual embeddings. Images are a rich, language-agnostic medium that can provide a \emph{bridge} across languages. For example, the English word ``cat'' might be found on webpages containing images of cats. Similarly, the German word ``katze'' (meaning cat) is likely to be found on other webpages containing similar (or perhaps identical) images of cats. Thus, images can be used to learn that these words have similar semantic content. Importantly, image-text data is generally available on the internet even for low-resource languages.

As image data has proliferated on the internet, tools for understanding images have advanced considerably. Convolutional neural networks (CNNs) have achieved roughly human-level or better performance on vision tasks, particularly classification \cite{imagenet, googlenet, Resnet}. During classification of an image, CNNs compute intermediate outputs that have been used as generic image features that perform well across a variety of vision tasks \cite{sharifoffshelf}. We use these image features to enforce that words associated with similar images have similar embeddings. Since words associated with similar images are likely to have similar semantic content, even across languages, our learned embeddings capture crosslingual similarity.

There has been other recent work on reducing the amount of supervision required to learn multilingual embeddings (cf.~Section~\ref{Related}). These methods take monolingual embeddings learned using existing methods and align them post-hoc in a shared embedding space. A limitation with post-hoc alignment of monolingual embeddings, first noticed by \citet{duong2017multilingual}, is that doing training of monolingual embeddings and alignment separately may lead to worse results than joint training of embeddings in one step. Since the monolingual embedding objective is distinct from the multilingual embedding objective, monolingual embeddings are not required to capture all information helpful for post-hoc multilingual alignment. Post-hoc alignment loses out on some information, whereas joint training does not. \citet{duong2017multilingual} observe improved results using a joint training method compared to a similar post-hoc method. Thus, a joint training approach is desirable.
To our knowledge, no previous method jointly learns multilingual word embeddings using weakly-supervised data available for low-resource languages.

To summarize: In this paper we propose an approach for learning multilingual word embeddings using image-text data jointly across all languages. We demonstrate that even a bag-of-words based embedding approach achieves performance competitive with the state-of-the-art on crosslingual semantic similarity tasks. We present experiments for understanding the effect of using pixel data as compared to co-occurrences alone. We also provide a method for training and making predictions on multilingual word embeddings even when the language of the text is unknown.

\section{Related Work}
\label{Related}

Most work on producing multilingual embeddings has relied on crosslingual human-labeled data, such as bilingual lexicons \cite{MikolovMapping, Ammar, Faruqui,xing2015normalized} or parallel/aligned corpora \cite{klementiev2012inducing, Ammar,luong2015bilingual,vulic2015bilingual}. These works are also largely bilingual due to either limitations of methods or the requirement for data that exists only for a few language pairs. Bilingual embeddings are less desirable because they do not leverage the relevant resources of other languages. For example, in learning bilingual embeddings for English and French, it may be useful to leverage resources in Spanish, since French and Spanish are closely related. Bilingual embeddings are also limited in their applications to just one language pair.

For instance, \citet{luong2015bilingual} propose BiSkip, a model that extends the skip-gram approach of \citet{skipgram} to a bilingual parallel corpus. The embedding for a word is trained to predict not only its own context, but also the contexts for corresponding words in a second corpus in a different language. \citet{Ammar} extend this approach further to multiple languages. This method, called MultiSkip, is compared to our methods in Section \ref{Results}.

\begin{figure*}
    \centering
    \includegraphics[width=.6\textwidth]{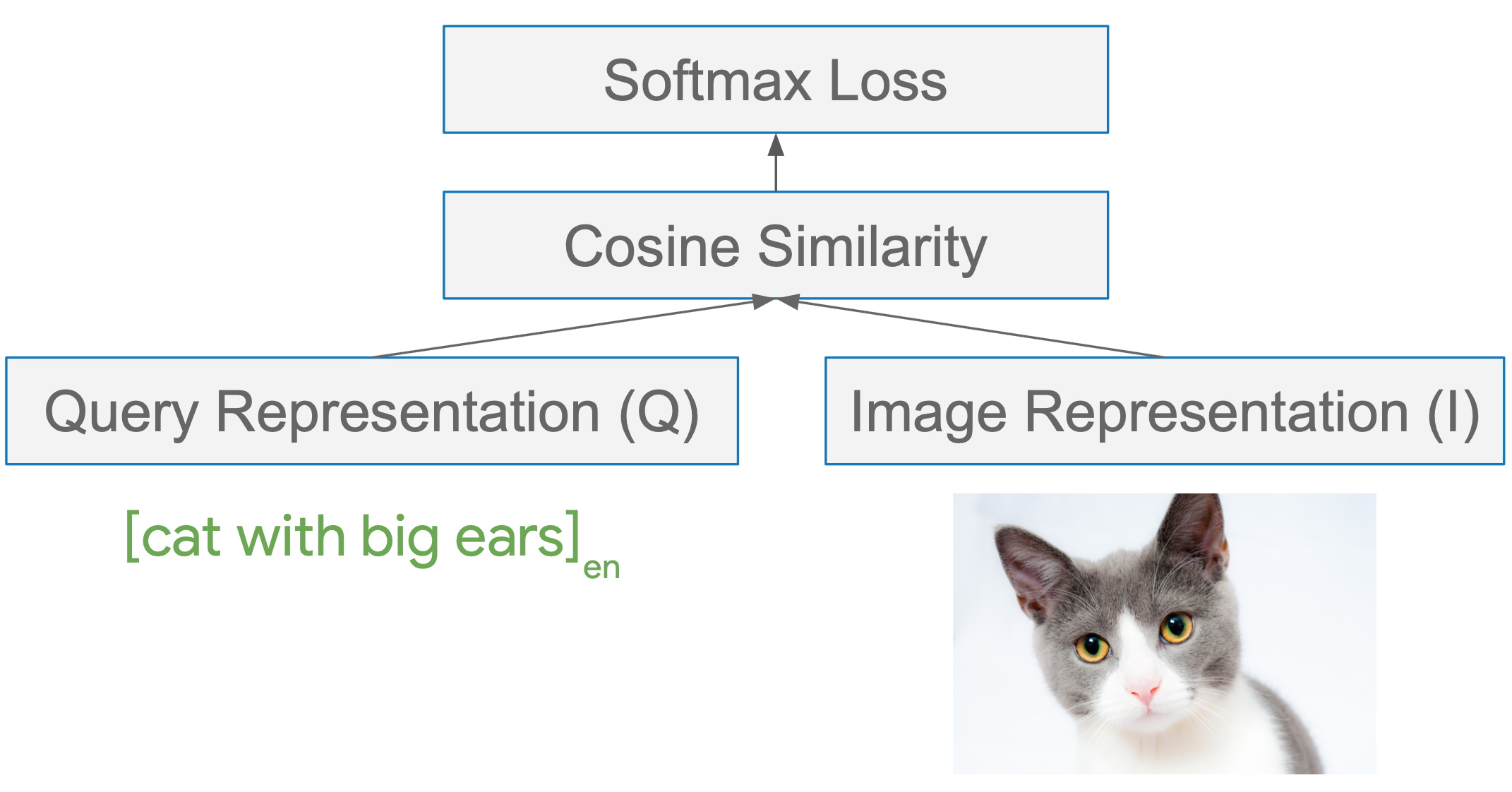}
    \caption{Our high-level approach for constraining query and image representations to be similar. The English query ``cat with big ears'' is mapped to $Q$, while the corresponding image example is mapped to $I$. We use the cosine similarity of these representations as input to a softmax loss function. The model task can be understood as predicting if an image is relevant to a given query.}
    \label{fig:highlevelapproach}
\end{figure*}

There has been some recent work on reducing the amount of human-labeled data required to learn multilingual embeddings, enabling work on low-resource languages \cite{smith2017offline,artetxe2017learning,MUSE}. These methods take monolingual embeddings learned using existing methods and align them post-hoc in a shared embedding space, exploiting the structural similarity of monolingual embedding spaces first noticed by \citet{MikolovMapping}. As discussed in Section \ref{Introduction}, post-hoc alignment of monolingual embeddings is inherently suboptimal. For example, \citet{smith2017offline} and \citet{artetxe2017learning} use human-labeled data, along with shared surface forms across languages, to learn an alignment in the bilingual setting. \citet{MUSE} build on this for the multilingual setting, using no human-labeled data and instead using an adversarial approach to maximize alignment between monolingual embedding spaces given their structural similarities. This method (MUSE) outperforms previous approaches and represents the state-of-the-art. We compare it to our methods in Section \ref{Results}. \looseness=-1

There has been other work using image-text data to improve image and caption representations for image tasks and to learn word translations \cite{karpathy2015deep, frome2013devise, GellaPivoting, Calixto, hewitt2018learning}, but no work using images to learn competitive multilingual word-level embeddings.

\section{Data}
We experiment using a dataset derived from Google Images search results\footnote{https://images.google.com}. The dataset consists of queries and the corresponding image search results. For example, one (query, image) pair might be ``cat with big ears'' and an image of a cat. Each (query, image) pair also has a weight corresponding to a relevance score of the image for the query. The dataset includes 3 billion (query, image, weight) triples, with 900 million unique images and 220 million unique queries. The data was prepared by first taking the query-image set, filtering to remove any personally identifiable information and adult content, and tokenizing the remaining queries by replacing special characters with spaces and trimming extraneous whitespace. Rare tokens (those that do not appear in queries at least six times) are filtered out. Each token in each query is given a language tag based on the user-set home language of the user making the search on Google Images. For example, if the query ``back pain'' is made by a user with English as her home language, then the query is stored as ``en:back en:pain''. The dataset includes queries in about 130 languages.

Though the specific dataset we use is proprietary, \citet{hewitt2018learning} have obtained a similar dataset, using the Google Images search interface, that comprises queries in 100 languages.

\section{Methods}

We present a series of experiments to investigate the usefulness of multimodal image-text data in learning multilingual embeddings. The crux of our method involves enforcing that for each query-image pair, the query representation ($Q$) is similar to the image representation ($I$). The query representation is a function of the word embeddings for each word in a (language-tagged) query, so enforcing this constraint on the query representation also has the effect of constraining the corresponding multilingual word embeddings.

Given some $Q$ and some $I$, we enforce that the representations are similar by maximizing their cosine similarity. We use a combination of cosine similarity and softmax objective to produce our loss. This high-level approach is illustrated in Figure \ref{fig:highlevelapproach}. In particular, we calculate unweighted loss as follows for a query $q$ and a corresponding image $i$: 
$$\textrm{loss}(\textrm{Query} \: q, \textrm{Image} \: i) = -\log {\frac{e^{\frac{Q_q^T I_i}{|Q_q| |I_i|}}}{\sum_{j} e^{\frac{Q_q^T I_j}{{|Q_q| |I_j|}}}}}$$
where $Q_q$ is the query representation for query $q$; $I_i$ is the image representation corresponding to image $i$; $j$ ranges over all images in the corpus; and $Q_q^T I_i$ is the dot product of the vectors $Q_q$ and $I_i$. Note that this requires that $Q_q$ and $I_j$ have identical dimensionality. If a weight $w$ is provided for the (query, image) pair, the loss is multiplied by the weight. Observe that $Q$ and $I$ remain unspecified for now: we detail different experiments involving different representations below.

In practice, given the size of our dataset, calculating the full denominator of the loss for a query, image pair would involve iterating through each image for each query, which is $O(n^2)$ in the number of training examples. To remedy this, we calculated the loss within each batch separately. That is, the denominator of the loss only involved summing over images in the same batch as the query. We used a batch size of 1000 for all experiments. In principle, the negative sampling approach used by \citet{word2vec} could be used instead to prevent quadratic time complexity.

\begin{figure*}
    \centering
    \includegraphics[width=.6\textwidth]{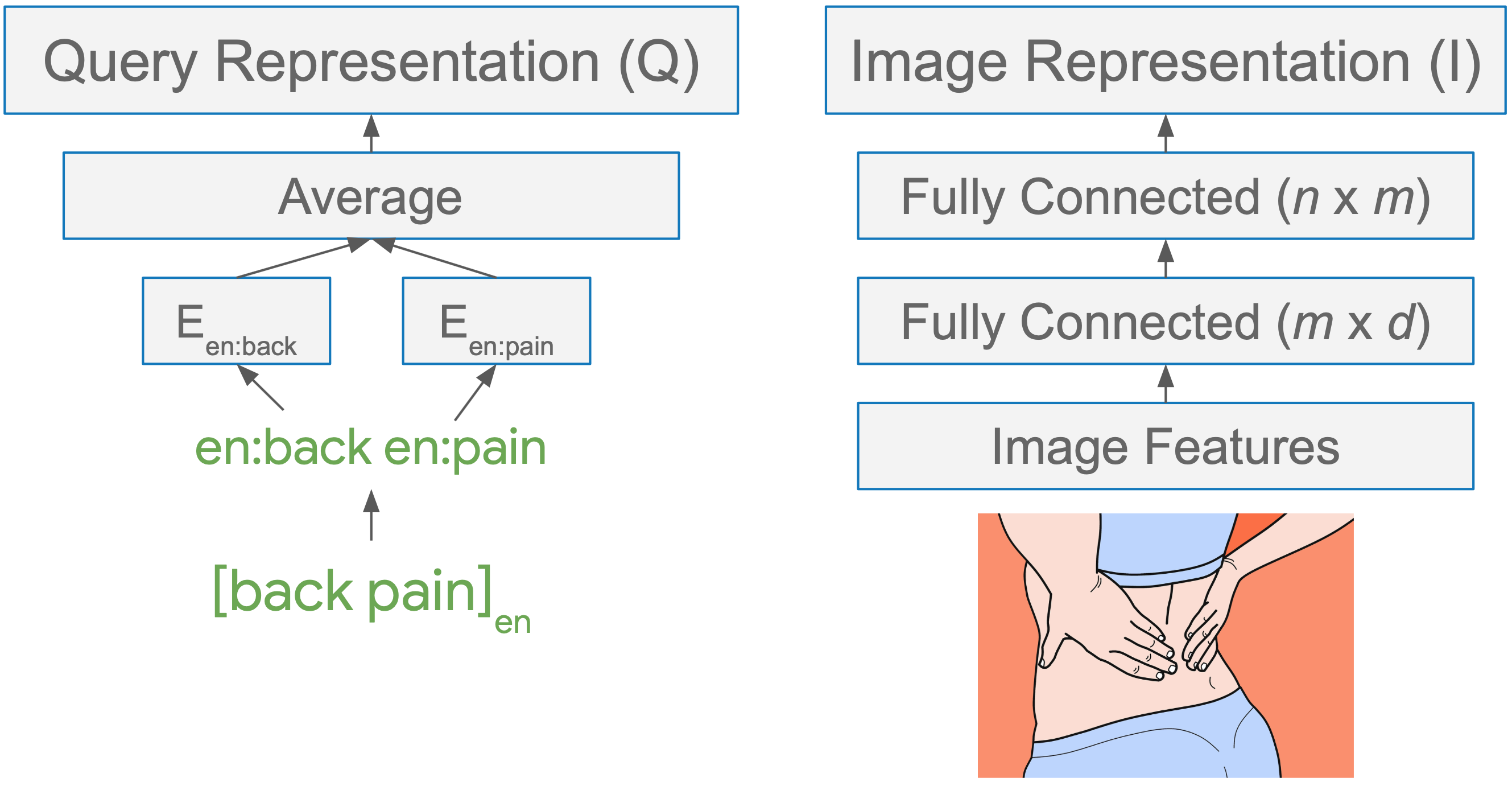}
    \caption{Our first method for calculating query and image representations, as presented in Section \ref{iuapproach}. To calculate the query representation, the multilingual embeddings for each language-prefixed token are averaged. To calculate the image representation, $d$-dimensional generic image features are passed through two fully-connected layers with $m$ and $n$ neurons.}
    \label{fig:leveragingimage}
\end{figure*}

We can interpret this loss function as producing a softmax classification task for queries and images: given a query, the model needs to predict the image relevant to that query. The cosine similarity between the image representation $I_i$ and the query representation $Q_q$ is normalized under softmax to produce a ``belief'' that the image $i$ is the image relevant to the query $q$. This is analogous to the skip-gram model proposed by \citet{skipgram}, although we use cosine similarity instead of dot product. Just as the skip-gram model ensures the embeddings of words are predictive of their contexts, our model ensures the embeddings of queries (and their constituent words) are predictive of images relevant to them. 

\subsection{Leveraging Image Understanding}
\label{iuapproach}

Given the natural co-occurrence of images and text on the internet and the availability of powerful generic features, a first approach is to use generic image features as the foundation for the image representation $I$. We apply two fully-connected layers to learn a transformation from image features to the final representation. We can compute the image representation $I_i$ for image $i$ as: 
$$I_i = ReLU(U * ReLU(Vf_i + b_1) + b_2)$$
where $f_i$ is a $d$-dimensional column vector representing generic image features for image $i$, $V$ is a $m \times d$ matrix, $b_1$ is an $m$-dimensional column vector, $U$ is a $n \times m$ matrix, and $b_2$ is an $n$-dimensional column vector. We use a rectified linear unit activation function after each fully-connected layer. 

We use 64-dimensional image features derived from image-text data using an approach similar to that used by \citet{juan2019graph}, who train image features to discriminate between fine-grained semantic image labels. We run two experiments with $m$ and $n$: in the first, $m = 200$ and $n = 100$ (producing 100-dimensional embeddings), and in the second, $m = 300$ and $n = 300$ (producing 300-dimensional embeddings).

For the query representation, we use a simple approach. The query representation is just the average of its constituent multilingual embeddings. Then, as the query representation is constrained to be similar to corresponding image representations, the multilingual embeddings (randomly initialized) are also constrained.

Note that each word in each query is prefixed with the language of the query. For example, the English query ``back pain'' is treated as ``en:back en:pain'', and the multilingual embeddings that are averaged are those for ``en:back'' and ``en:pain''. This means that words in different languages with shared surface forms are given separate embeddings. We experiment with shared embeddings for words with shared surface forms in Section \ref{langunaware}.

In practice, we use a fixed multilingual vocabulary for the word embeddings, given the size of the dataset. Out-of-vocabulary words are handled by hashing them to a fixed number of embedding buckets (we use 1,000,000). That is, there are 1,000,000 embeddings for all out-of-vocabulary words, and the assignment of embedding for each word is determined by a hash function.

Our approach for leveraging image understanding is shown in Figure \ref{fig:leveragingimage}.

\subsection{Co-Occurrence Only}
\label{coocs}


Another approach for generating query and image representations is treating images as a black box. Without using pixel data, how well can we do? Given the statistics of our dataset (3B query, image pairs with 220M unique queries and 900M unique images), we know that different queries co-occur with the same images. Intuitively, if a query $q_1$ co-occurs with many of the same images as query $q_2$, then $q_1$ and $q_2$ are likely to be semantically similar, regardless of the visual content of the shared images. Thus, we can use a method that uses only co-occurrence statistics to better understand how well we can capture relationships between queries. This method serves as a baseline to our initial approach leveraging image understanding.

In this setting, we keep query representations the same, and we modify image representations as follows: the image representation for an image is a randomly initialized, trainable vector (of the same dimensionality as the query representation, to ensure the cosine similarity can be calculated). The intuition for this approach is that if two queries are both associated with an image, their query representations will both be constrained to be similar to the same vector, and so the query representations themselves are constrained to be similar. This approach
is a simple way to adapt our method to make use of only co-occurrence statistics.

One concern with this approach is that many queries may not have significant image co-occurrences with other queries. In particular, there are likely many images associated with only a single query. These isolated images pull query representations toward their respective random image representations (adding noise), but do not provide any information about the relationships between queries. Additionally, even for images associated with multiple queries, if these queries are all within language, then they may not be very helpful for learning multilingual embeddings. Consequently, we run two experiments: one with the original dataset and one with a subset of the dataset that contains only images associated with queries in at least two different languages. This subset of the dataset has 540 million query, image pairs (down from 3 billion). For both experiments, we use $m = 200$ and $n = 100$ and produce 100-dimensional embeddings.

\subsection{Language Unaware Query Representation}
\label{langunaware}

In Section \ref{iuapproach}, our method for computing query representations involved prepending language prefixes to each token, ensuring that the multilingual embedding for the English word ``pain'' is distinct from that for the French word ``pain'' (meaning bread). These query representations are \emph{language aware}, meaning that a language tag is required for each query during both training and prediction. In the weakly-supervised setting, we may want to relax this requirement, as language-tagged data is not always readily available.

\begin{figure}[t]
    \centering
    \includegraphics[width=.32\textwidth]{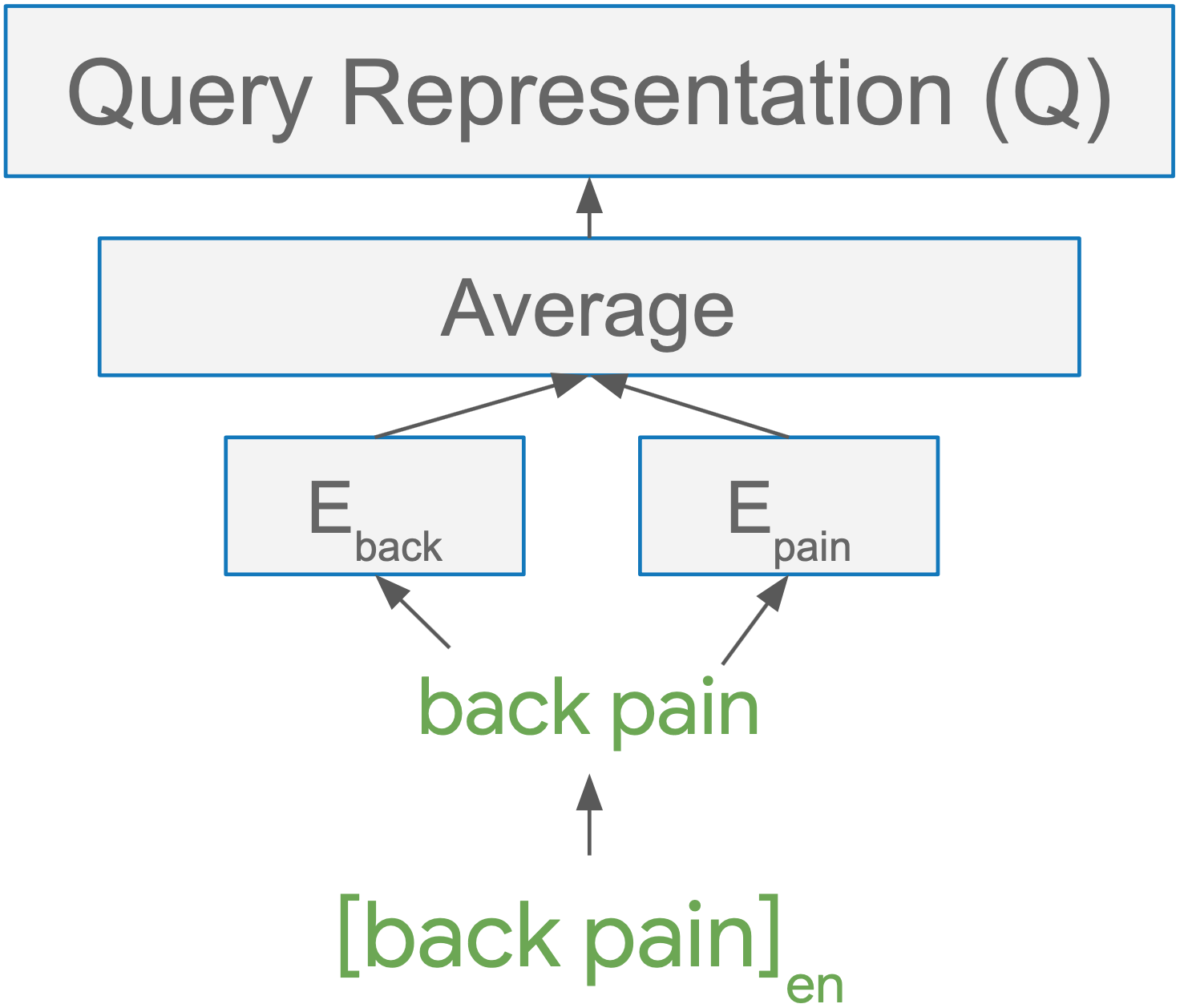}
    \caption{In our language unaware approach, language tags are not prepended to each token, so the word ``pain'' in English and French share an embedding.}
    \label{fig:languageunawareapproach}
\end{figure}

In our language unaware setting, language tags are not necessary. Each surface form in each query has a distinct embedding, and words with shared surface forms across languages (e.g., English ``pain'' and French ``pain'') have a shared embedding. In this sense, shared surface forms are used as a bridge between languages. This is illustrated in Figure \ref{fig:languageunawareapproach}. This may be helpful in certain cases, as for English  ``actor'' and Spanish ``actor''. The image representations leverage generic image features, exactly as in Section \ref{iuapproach}. In our language-unaware experiment, we use $m = 200$ and $n = 100$ and produce 100-dimensional embeddings.

\subsection{Evaluation}
We evaluate our learned multilingual embeddings using six crosslingual semantic similarity tasks, two multilingual document classification tasks, and 13 monolingual semantic similarity tasks. We adapt code from \citet{Ammar} and \citet{faruqui-2014:SystemDemo} for evaluation.

\paragraph{Crosslingual Semantic Similarity}
\label{sec:crosslingualsemanticsimilarity}
This task measures how well multilingual embeddings capture semantic similarity of words, as judged by human raters. The task consists of a series of crosslingual word pairs. For each word pair in the task, human raters judge how semantically similar the words are. The model also predicts how similar the words are, using the cosine similarity between the embeddings. The score on the task is the Spearman correlation between the human ratings and the model predictions. 

The specific six subtasks we use are part of the Rubenstein-Goodenough dataset \cite{rubenstein1965contextual} and detailed by \citet{Ammar}. We also include an additional task aggregating the six subtasks.

\paragraph{Multilingual Document Classification}
\label{evaluationclassification}
In this task, a classifier built on top of learned multilingual embeddings is trained on the RCV corpus of newswire text as in \citet{klementiev2012inducing} and \citet{Ammar}. The corpus consists of documents in seven languages on four topics, and the classifier predicts the topic. The score on the task is test accuracy. Note that each document is monolingual, so this task measures performance within languages for multiple languages (as opposed to crosslingual performance).

\paragraph{Monolingual Semantic Similarity}
This task is the same as the crosslingual semantic similarity task described above, but all word pairs are in English. We use this to understand how monolingual performance differs across methods. We present an average score across the 13 subtasks provided by \citet{faruqui-2014:SystemDemo}.

\paragraph{Coverage}
Evaluation tasks also report a coverage, which is the fraction of the test data that a set of multilingual embeddings is able to make predictions on. This is needed because not every word in the evaluation task has a corresponding learned multilingual embedding. Thus, if coverage is low, scores are less likely to be reliable.

\section{Results and Conclusions}
\label{Results}

\begin{table*}[t]
\centering
\begingroup
\renewcommand*{\arraystretch}{1.0}

\begin{tabular}{llllllll}
\toprule
                          & en+es                                       & en+de                                       & en+fr                                       & de+es                                        & de+fr                                       & fr+es                                       & all                                         \\
                          \midrule
ImageVec 100-Dim          & .75 {\scriptsize[.87]} & .77 {\scriptsize[.87]} & .84 {\scriptsize[.74]} & .80 {\scriptsize[.83]}  & .76 {\scriptsize[.77]} & .77 {\scriptsize[.73]} & .79 {\scriptsize[.81]} \\
ImageVec 300-Dim          & {\bf .79} {\scriptsize[.87]} & .81 {\scriptsize[.87]} & {\bf .86} {\scriptsize[.74]} & .81 {\scriptsize[.83]}  & {\bf .77} {\scriptsize[.77]} & {\bf .80} {\scriptsize[.73]} & {\bf .82} {\scriptsize[.81]} \\
ImageVec Baseline         & .10 {\scriptsize[.87]} & .03 {\scriptsize[.87]} & .14 {\scriptsize[.74]} & -.25 {\scriptsize[.83]} & .07 {\scriptsize[.77]} & .15 {\scriptsize[.73]} & .08 {\scriptsize[.81]} \\
ImageVec Baseline 2 Lang. & .27 {\scriptsize[.87]} & .38 {\scriptsize[.79]} & .23 {\scriptsize[.74]} & .26 {\scriptsize[.75]}  & .16 {\scriptsize[.75]} & .27 {\scriptsize[.73]} & .28 {\scriptsize[.78]} \\
ImageVec Lang. Unaware    & .59 {\scriptsize[.87]} & .62 {\scriptsize[.87]} & .79 {\scriptsize[.74]} & .63 {\scriptsize[.83]}  & .73 {\scriptsize[.77]} & .73 {\scriptsize[.73]} & .67 {\scriptsize[.81]} \\
\midrule
multiSkip 40-Dim          & .51 {\scriptsize[.83]} & .67 {\scriptsize[.75]} & .44 {\scriptsize[.70]} & .39 {\scriptsize[.63]}  & .29 {\scriptsize[.56]} & .43 {\scriptsize[.60]} & .49 {\scriptsize[.68]} \\
multiSkip 512-Dim         & .43 {\scriptsize[.83]} & .73 {\scriptsize[.76]} & .62 {\scriptsize[.70]} & .43 {\scriptsize[.63]}  & .24 {\scriptsize[.56]} & .48 {\scriptsize[.60]} & .50 {\scriptsize[.69]} \\
MUSE 300-Dim              & .76 {\scriptsize[.87]} & {\bf .85} {\scriptsize[.86]} & .79 {\scriptsize[.74]} & {\bf .83} {\scriptsize[.81]}  & .73 {\scriptsize[.77]} & .74 {\scriptsize[.73]} & .79 {\scriptsize[.80]}\\
\bottomrule
\end{tabular}
\endgroup
\caption{Crosslingual semantic similarity scores (Spearman's $\rho$) across six subtasks for ImageVec (our method) and previous work. Coverage is in brackets. The last column indicates the combined score across all subtasks. Best scores on each subtask are bolded.}
    \label{tab:crosslingual}
\end{table*}

\begin{table}[t]
\centering
\begingroup
\renewcommand*{\arraystretch}{1.0}

\begin{tabular}{lll}
\toprule
                          & {\small en+da+it}                                    & {\small 7 Lang.}                        \\
                          \midrule
ImageVec 100-Dim          & .74 {\scriptsize[.60]} & .79 {\scriptsize[.52]} \\
ImageVec 300-Dim          & .80 {\scriptsize[.60]} & .84 {\scriptsize[.52]} \\
ImageVec Baseline         & .60 {\scriptsize[.60]} & .59 {\scriptsize[.52]} \\
ImageVec Baseline 2 Lang. & .65 {\scriptsize[.45]} & .65 {\scriptsize[.36]} \\
ImageVec Lang. Unaware    & .73 {\scriptsize[.60]} & .78 {\scriptsize[.52]} \\
\midrule
multiSkip 40-Dim          & .77 {\scriptsize[.45]} & .82 {\scriptsize[.44]} \\
multiSkip 512-Dim         & .87 {\scriptsize[.48]} & .91 {\scriptsize[.46]} \\
MUSE 300-Dim              & {\bf .87} {\scriptsize[.54]} & {\bf .91} {\scriptsize[.51]}\\
\bottomrule
\end{tabular}
\endgroup
\caption{Multilingual document classification accuracy scores across two subtasks for ImageVec (our method) and previous work. Coverage is in brackets. Best scores are bolded (ties broken by coverage).}
    \label{tab:classification}
\end{table}

We first present results on the crosslingual semantic similarity and multilingual document classification for our previously described experiments. We compare against the multiSkip approach by \citet{Ammar} and the state-of-the-art MUSE approach by \citet{MUSE}. Results for crosslingual semantic similarity are presented in Table \ref{tab:crosslingual}, and results for multilingual document classification are presented in Table \ref{tab:classification}. 

Our experiments corresponding to Section \ref{iuapproach} are titled \emph{ImageVec 100-Dim} and \emph{ImageVec 300-Dim} in Tables \ref{tab:crosslingual} and \ref{tab:classification}. Both experiments significantly outperform the multiSkip experiments in all crosslingual semantic similarity subtasks, and the 300-dimensional experiment slightly outperforms MUSE as well. Note that coverage scores are generally around 0.8 for these experiments. In multilingual document classification, MUSE achieves the best scores, and while our 300-dimensional experiment outperforms the multiSkip 40-dimensional experiment, it does not perform as well as the 512-dimensional experiment. Note that coverage scores are lower on these tasks. 

One possible explanation for the difference in performance across the crosslingual semantic similarity task and multilingual document classification task is that the former measures crosslingual performance, whereas the latter measures monolingual performance in multiple languages, as described in Section \ref{evaluationclassification}. We briefly discuss further evidence that our models perform less well in the monolingual context below.

\paragraph{Is Image Understanding Necessary?}

Comparing the experiments leveraging image understanding to our co-occurrence-only baseline experiments \emph{ImageVec Baseline} and \emph{ImageVec Baseline 2 Lang} described in Section \ref{coocs}, we see that performance is significantly degraded without pixel data (note that both experiments use 100-dimensional embeddings). Still, the results for multilingual document classification, in particular, show that we are able to learn multilingual word embeddings using co-occurrence between queries and images alone. 

Interestingly, we can see that performance in the experiment in which images are filtered to be associated with at least two languages appears better than the baseline experiment on the full dataset (although coverage is low for multilingual document classification). As mentioned in Section \ref{coocs}, this may be because images without multiple queries degrade performance by introducing noise to the optimization problem. We also experimented with the same filtering on the experiments using image understanding to see if this could further boost performance (results not shown), but this reduced performance to a similar extent as random data filtering. This is likely because even isolated images (with just one query associated with an image) are still helpful for the task in this case, since the use of generic image features still constrains queries associated with similar images to have similar representations. 

Even in the filtered baseline, results for both tasks are significantly lower than the methods leveraging image understanding, indicating that while co-occurrence data alone is useful, pixel data may be needed to learn competitive multilingual embeddings using our method.

\paragraph{Language Unaware Learning}

The language unaware setting only differs from the language aware one when words share a common surface form. In some cases, words sharing a common surface form have the same meaning across languages (\emph{i.e.,} cognates). An example is ``actor'' in English and Spanish. In these cases, the language unaware setting may boost performance, as the embedding for ``actor'' effectively has more training data behind it. In other cases, words sharing a common surface form have different meanings across languages (\emph{i.e.,} false cognates). An example is ``pain'' in English and French. In these cases, we expect language unawareness to reduce performance, especially if the meanings of false cognates are very different.

Our results for our 100-dimensional language unaware embeddings are presented in Tables \ref{tab:crosslingual} and \ref{tab:classification} as \emph{ImageVec Lang. Unaware}. We can see that this experiment performs worse on crosslingual semantic similarity but about the same on multilingual document classification as the 100-dimensional language aware experiment (\emph{ImageVec 100-Dim}). Still, on crosslingual semantic similarity, it significantly outperforms both multiSkip experiments. Thus, in applications where language unaware training or prediction is important, our method produces multilingual embeddings competitive with other language aware approaches.

\paragraph{Effect of Embedding Size}
In these experiments, embeddings with higher dimensionalities generally perform better in both evaluation tasks. 300-dimensional embeddings produced using our method slightly outperform 100-dimensional ones in every subtask for both tasks.

\begin{table}[t]
\centering
\begingroup
\renewcommand*{\arraystretch}{1.0}

\begin{tabular}{ll}
\toprule
                          & {\small avg. score}                         \\
                          \midrule
ImageVec 100-Dim          & .48 {\scriptsize[.98]} \\
ImageVec 300-Dim          & .48 {\scriptsize[.98]} \\
ImageVec Baseline         & .24 {\scriptsize[.98]} \\
ImageVec Baseline 2 Lang. & .33 {\scriptsize[.95]} \\
ImageVec Lang. Unaware    & .42 {\scriptsize[.98]} \\
\midrule
multiSkip 40-Dim          & .44 {\scriptsize[.94]} \\
multiSkip 512-Dim         & .44 {\scriptsize[.96]} \\
MUSE 300-Dim              & {\bf .62} {\scriptsize[.97]} \\
\bottomrule
\end{tabular}
\endgroup
\caption{Average monolingual semantic similarity score (Spearman's $\rho$) across 13 subtasks for ImageVec (our method) and previous work. Average coverage is in brackets. Best score is bolded.}
    \label{tab:monolingual}
\end{table}

\paragraph{Monolingual Embedding Quality}
\label{sec:monolingualquality}
As mentioned earlier in Section \ref{Results}, we suspect that the difference in performance (as compared to MUSE) on crosslingual semantic similarity and multilingual document classification for our experiments might be due to reduced monolingual performance. After all, other methods train by leveraging word contexts (and subword information, in the case of MUSE) in a large monolingual corpus, whereas we use only images as a bridge between words within and across languages. Especially for words representing abstract concepts without obvious image associations (consider the word ``democracy''), it is likely that our method would produce lower quality within-language embeddings than text-only methods. This is not unexpected: \citet{hewitt2018learning} found that word translations learned via images are worse for more abstract words and \citet{kiela2014improving} found that using image data is unhelpful for improving the quality of representations for some concepts.

It stands to reason then that our method would produce weaker monolingual performance. To test this, we ran 13 English monolingual semantic similarity tasks on each experiment. We present average scores in Table \ref{tab:monolingual}. We can see that 300-dimensional embeddings produced using our method fare significantly worse than MUSE embeddings, although they perform similarly to the multiSkip embeddings. For comparison, competitive English word embeddings achieve results similar to MUSE. This suggests that there is significant room for improvement within language (at least for English) in the quality of our learned multilingual embeddings. Improving monolingual performance would also likely boost scores across other tasks, motivating future work in this direction.

\section{Discussion}
\label{Discussion}
We demonstrated how to learn competitive multilingual word embeddings using image-text data -- which is available for low-resource languages. We have presented experiments for understanding the effect of using pixel data as compared to co-occurrences alone. We have also proposed a method for training and making predictions on multilingual word embeddings even when language tags for words are unavailable. Using a simple bag-of-words approach, we achieve performance competitive with the state-of-the-art on crosslingual semantic similarity tasks.

We have also identified a direction for future work: within language performance is weaker than the state-of-the-art, likely because our work leveraged only image-text data rather than a large monolingual corpus. Fortunately, our joint training approach provides a simple extension of our method for future work: multi-task joint training. For example, in a triple-task setting, we can simultaneously {\bf (1)} constrain query and relevant image representations to be similar and {\bf (2)} constrain word embeddings to be predictive of context in large monolingual corpora and {\bf (3)} constrain representations for parallel text across languages to be similar. For the second task, implementing recent advances in producing monolingual embeddings, such as using subword information, is likely to improve results. Multilingual embeddings learned in a multi-task setting would reap both the benefits of our methods and existing methods for producing word embeddings. For example, while our method is likely to perform worse for more abstract words, when combined with existing approaches it is likely to achieve more consistent performance.

An interesting effect of our approach is that queries and images are embedded into a shared space through the query and image representations. This setup enables a range of future research directions and applications, including better image features, better monolingual text representations (especially for visual tasks), nearest-neighbor search for text or images given one modality (or both), and joint prediction using text and images. 

\paragraph*{Acknowledgments}
We thank Tom Duerig, Evgeniy Gabrilovich, Dan Gillick, Raphael Hoffmann, Zhen Li, Alessandro Presta, 
Aleksei Timofeev, Radu Soricut, and our reviewers for their insightful feedback and comments.

\bibliography{naaclhlt2019}
\bibliographystyle{acl_natbib}



\end{document}